\documentclass[sigconf]{acmart}
\setcopyright{none}
\renewcommand\footnotetextcopyrightpermission[1]{}
\acmConference{}{}{}

\usepackage{multirow}
\usepackage{bm}
\AtBeginDocument{%
  }

\makeatletter

\renewcommand*{\@fnsymbol}[1]{%
  \ensuremath{\ifcase#1\or \dagger\or \ast\or \ddagger\or
     \mathsection\or \mathparagraph\or \|\or **\or \dagger\dagger
     \or \ddagger\ddagger \else\@ctrerr\fi}}
\makeatother

\begin{document}

\title{Beyond Feature and Structure Alignment: Learning Transferable Propagation Knowledge for Graph Foundation Models}

\author{Yi Wang}
\authornote{Both authors contributed equally to this research.}
\email{wangyi076@tju.edu.cn}
\affiliation{
  \institution{Tianjin University}
  \city{Tianjin}
  \country{China}
}

\author{Jitao Zhao}
\authornotemark[1]
\email{zjtao@tju.edu.cn}
\affiliation{
  \institution{Tianjin University}
  \city{Tianjin}
  \country{China}
}

\author{Di Jin}
\email{jindi@tju.edu.cn}
\affiliation{%
  \institution{Tianjin University}
  \city{Tianjin}
  \country{China}
}

\author{Dongxiao He}
\authornote{Corresponding author.}
\email{hedongxiao@tju.edu.cn}
\affiliation{%
  \institution{Tianjin University}
  \city{Tianjin}
  \country{China}
}

\renewcommand{\shortauthors}{Yi Wang et al.}

\begin{abstract}
Graph Foundation Models (GFMs) have recently emerged as a promising paradigm for enabling knowledge transfer across diverse domains. Unlike traditional graph learning methods that are typically designed for in-domain settings, GFMs aim to learn transferable knowledge that can generalize to unseen graph domains.
However, unlike language or visual data, graphs lack intrinsic and unified representation units, such as tokens in language and patches in vision, making it challenging to identify transferable knowledge units for building graph foundation models. Existing graph foundation models mainly focus on mitigating domain discrepancies through feature alignment and structure alignment, while overlooking the exploration of transferable knowledge units underlying graph data. Moreover, these methods generally rely on fixed propagation mechanisms during message passing, overlooking the heterogeneity in propagation patterns, as different edges may exhibit distinct propagation patterns for different feature dimensions. To address these limitations, we propose a Propagation-aware Graph Foundation Model (ProGFM), which regards the propagation relationships between edges and feature dimensions as transferable knowledge units. Through a propagation relationship prototype bank, ProGFM learns cross-domain transferable propagation knowledge, enabling adaptive information aggregation in unseen graph domains. Extensive experiments across various cross-domain transfer scenarios demonstrate that ProGFM possesses strong cross-domain knowledge transfer capability and exhibits superior generalization performance compared with existing methods.

\end{abstract}

\keywords{Graph Neural Networks, Graph Representation Learning, Graph Foundation Models}

\maketitle

\section{Introduction}
Graphs, as a powerful data form for modeling complex relationships in the real world, have been widely applied across various domains, such as social networks \cite{social_networks_1, social_networks_2}, recommender systems \cite{recommendation_systems_1, recommendation_systems_2}, and transportation networks \cite{transportation_1, transportation_2}. To effectively mine the information contained in the graph data, numerous Graph Neural Networks (GNNs), e.g., GCN \cite{GCN} and GAT \cite{GAT}, have been proposed. However, these methods are typically designed for in-domain settings, and often require rebuilding when applied to new graph domains \cite{GFMServeyShiChuan}. With the remarkable success of foundation models in natural language and computer vision, Graph Foundation Models (GFMs) have recently attracted increasing attention \cite{GFM_increasing_attention}. GFMs aim to learn transferable graph knowledge from multi-domain graph data, enabling effective generalization to unseen graph domains \cite{BRIDGE}.

Unlike language or visual data, graphs lack intrinsic and unified representation units, such as tokens in language and patches in vision \cite{REEF, All_in}. Graphs are typically derived from abstract modeling of complex relational systems in the real world, where nodes, edges, and features often exhibit diverse meanings across different domains \cite{OneforAll}. For example, user attributes in social networks, textual features in citation networks, and atomic attributes in molecular graphs have different domain-specific semantics. This fundamental difference makes it challenging to identify transferable knowledge units underlying graph data, which becomes a key obstacle for developing general graph foundation models.

Existing graph foundation models mainly achieve cross-domain generalization by aligning graph data from different domains. Feature alignment methods attempt to transform graph features from different domains into a shared representation space through feature mapping or feature reconstruction \cite{MUG, TIG}. Structure alignment methods reduce structural discrepancies through graph structure transformation or structure reconstruction \cite{MDGFM, TFSGFM}. However, these approaches mainly alleviate distribution shifts across graph domains, overlooking the exploration of transferable knowledge units underlying graph data, thereby limiting their ability to achieve effective cross-domain knowledge transfer.

Moreover, identifying transferable knowledge units underlying graph data needs to consider an important graph property, message propagation pattern, which is the fundamental mechanism for simultaneously embedding structure and features. Existing graph foundation models generally rely on fixed propagation mechanisms during message passing \cite{TIG, MDGPT}. In reality, due to the heterogeneity of semantics and relational patterns \cite{OOD_GFM}, different edges may exhibit distinct propagation patterns for different feature dimensions. Such fixed propagation mechanisms overlook the heterogeneity of propagation patterns and implicitly assume that different features can be propagated across edges in the same manner, limiting the expressive capacity of existing graph foundation models.

To address these limitations, the key step is to identify transferable knowledge units, which can represent truly transferable graph knowledge and model diverse propagation patterns.
We argue that propagation relationships between edges and feature dimensions have the potential to serve as such transferable knowledge units.
Unlike feature semantics, which vary significantly across graph domains and are difficult to directly transfer, propagation relationships between edges and feature dimensions exhibit stronger cross-domain transferability.
Specifically, across different graph domains, some edges may possess similar propagation relationships in certain feature dimensions (similar to the homophily at the feature dimension level in previous studies \cite{yangliang, TFSGFM}) and thus correspond to similar propagation patterns, even when these features carry completely different semantics.

Based on the above observations, we propose a Propagation-aware Graph Foundation Model (ProGFM), which treats propagation relationships between edges and feature dimensions as transferable knowledge units. Unlike existing methods that mainly achieve cross-domain adaptation through feature alignment \cite{TIG}, ProGFM does not assume that features at the same dimension across different domains share consistent semantics. Instead, it models the propagation relationship between each edge and each feature dimension based on the relative feature differences between connected nodes. By constructing a propagation relationship prototype bank, ProGFM captures cross-domain transferable propagation knowledge and enables adaptive information aggregation in unseen graph domains. In this way, ProGFM provides a new perspective for knowledge transfer in open graph environments by treating cross-domain shared propagation relationships as transferable knowledge units.

Our main contributions are summarized as follows:
\begin{itemize}
    \item We introduce a new perspective on knowledge transfer for graph foundation models. We reveal that propagation relationships between edges and features can serve as transferable knowledge units independent of specific domain semantics, providing a new direction for general graph modeling in open graph environments.

    \item We propose ProGFM, a graph foundation model framework that explicitly models propagation relationships between edges and features dimensions. Through a propagation relationship prototype bank, ProGFM can learn cross-domain transferable propagation knowledge and effectively generalize to unseen graph domains.

    \item We conduct extensive experiments across multiple cross-domain transfer scenarios to systematically evaluate the effectiveness and transferability of ProGFM. The results demonstrate that ProGFM achieves superior generalization performance on unseen graph domains compared with existing graph foundation models.

\end{itemize}

\section{Preliminaries}

In this section, we introduce the preliminary concepts and notations used throughout this paper.
Specifically, we first introduce the formulation of graph data and the concept of graph foundation model, followed by the definition of relative feature difference. 
The relative feature difference is defined to quantify the relative discrepancy between connected nodes along each feature dimension, which will be further used for modeling propagation relationships.

\subsection{Graph Data}
Formally, a graph can be represented as $G=(V, \mathcal{E}, \mathbf{A}, \mathbf{X}, \mathbf{Y})$, where $V=\{v_1,v_2,\dots,v_N\}$ denotes the node set and $\mathcal{E}$ denotes the edge set. $\mathbf{A} \in \mathbb{R}^{N \times N}$ represents the adjacency matrix, where $A_{ij}$ indicates whether nodes $v_i$ and $v_j$ are connected. $\mathbf{X} \in \mathbb{R}^{N \times d}$ denotes the node feature matrix, where $d$ denotes the feature dimensionality. $\mathbf{Y}=\{y_1,y_2,\dots,y_N\}$ denotes the node labels. For a node $v_i$, its feature vector is denoted as $\mathbf{x}_i = [x_{i,1},x_{i,2},\dots,x_{i,d}]$, where $x_{i,k}$ represents the value of the $k$-th feature dimension.

\subsection{Graph Foundation Model}
Given a collection of graphs from multiple domains, denoted as $\mathcal{G}_{pre} = \{G^{(1)},G^{(2)},\dots,G^{(M)}\}$, a graph foundation model aims to learn cross-domain transferable graph knowledge from these diverse pre-training graphs. For an unseen target graph domain $G^{(t)} \notin \mathcal{G}_{pre}$, the graph foundation model can achieve effective adaptation based on the knowledge learned from $\mathcal{G}_{pre}$.

\subsection{Relative Feature Difference}
\begin{definition}[Relative Feature Difference]
Given an edge $e_{ij} \in \mathcal{E}$ connecting nodes $v_i$ and $v_j$, we define the relative feature difference between the connected nodes on each feature dimension. Specifically, for the $k$-th feature dimension, the relative feature difference is calculated as:
\begin{equation} 
r_{ij,k} = \frac{\left|x_{i,k} - x_{j,k}\right|}{\text{range}_k},
\end{equation}
\begin{equation} 
\text{range}_k = \text{max}\left(\mathbf{X}_{:,k}\right) - \text{min}\left(\mathbf{X}_{:,k}\right) + \epsilon, 
\end{equation}
where $x_{i,k}$ and $x_{j,k}$ denote the values of the $k$-th feature dimension of nodes $v_i$ and $v_j$, respectively. $\mathbf{X}_{:,k}$ represents the $k$-th feature dimension of all nodes, and $\text{max}\left(\mathbf{X}_{:,k}\right)$ and $\text{min}\left(\mathbf{X}_{:,k}\right)$ denote its maximum and minimum values. $\epsilon$ is a small constant for numerical stability. A larger $r_{ij,k}$ indicates a larger relative difference between connected nodes on the $k$-th feature dimension.
\end{definition}

\begin{figure*}[htbp]
    \centering
    \includegraphics[width=\textwidth]{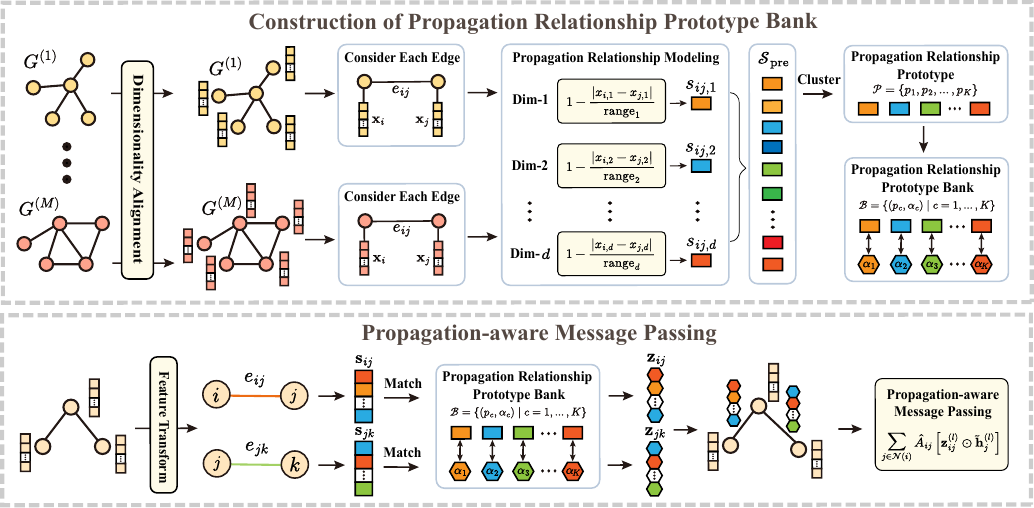}
    \caption{The overview of ProGFM.}
    \Description{The overview of ProGFM.}
    \label{fig:model}
\end{figure*}

\section{Method}
\subsection{Overview}
Unlike existing graph foundation models that mainly align node features or graph structures across domains, ProGFM focuses on modeling the propagation relationships between edges and feature dimensions as domain-agnostic knowledge units. Specifically, ProGFM consists of three main stages, as illustrated in Figure \ref{fig:model}. First, we align the feature dimensionality of graphs from different domains to provide a consistent input space for graph representation learning, without enforcing semantic alignment of domain-specific features. Second, based on the dimension-aligned features, we model the propagation relationships between each edge and each feature dimension according to the relative feature differences between connected nodes. The propagation relationships collected from multiple source domains are then clustered to construct a propagation relationship prototype bank, where each prototype captures a representative propagation relationship shared across domains. Each propagation relationship prototype is further associated with a learnable scalar parameter that represents its corresponding propagation strength. Finally, we introduce a propagation-aware message passing mechanism. Based on the propagation relationships between each edge and each feature dimension, each edge is associated with a propagation strength vector, which is used to modulate information aggregation of connected nodes.
\subsection{Feature Dimensionality Alignment}
Graphs from different domains usually contain node features with different dimensionalities \cite{GraphAny, TIG, AnyGraph}. Such dimensional discrepancies make it difficult for a graph model to directly process graphs from different domains. Therefore, we need to align the feature dimensionality of graphs from different domains into a shared dimensional space.

Specifically, for any graph $G^{(i)}=(V^{(i)}, \mathcal{E}^{(i)}, \mathbf{A}^{(i)}, \mathbf{X}^{(i)}, \mathbf{Y}^{(i)})$, we transform the original features into a unified dimensional space:
\begin{equation} 
\mathbf{\tilde{X}}^{(i)} = f_{\text{align}}(\mathbf{{X}}^{(i)}) \in \mathbb{R}^{N^{(i)} \times d},
\end{equation}
where $\mathbf{X}^{(i)} \in \mathbb{R}^{N^{(i)} \times d^{(i)}}$ denotes the original node feature matrix, $d^{(i)}$ represents the domain-specific feature dimensionality, $d$ denotes the unified feature dimensionality shared across different graph domains, and $f_{\text{align}}(\cdot)$ represents the feature dimensionality alignment function. The alignment function can be implemented using various projection or dimensionality reduction methods, such as Multi-Layer Perceptron, Singular Value Decomposition (SVD) \cite{svd}, or Principal Component Analysis \cite{pca}. In this work, we adopt SVD to perform feature dimensionality alignment.

By performing feature dimensionality alignment, node features from different graph domains are projected into a unified dimensional space, providing a unified input format for subsequent propagation relationship modeling and propagation-aware message passing.

\subsection{Construction of Propagation Relationship Prototype Bank}
Based on the dimension-aligned node features, the next question is what knowledge can be transferred across different graph domains. Although feature semantics may vary significantly across different graph domains, similar propagation relationships may still exist. Therefore, ProGFM models the propagation relationships between each edge and each feature dimension as transferable knowledge units.

In graph message passing, the relative feature difference between connected nodes on a particular dimension can reflect the propagation pattern of that feature dimension along the edge, and similar relative feature differences tend to indicate similar propagation patterns.
Therefore, we can model the propagation relationships between each edge and each feature dimension based on the relative feature differences between connected nodes. Specifically, following the definition in the preliminaries, $r_{ij,k}$ denotes the relative feature difference between nodes $v_i$ and $v_j$ on the $k$-th feature dimension. The propagation relationship between edge $e_{ij}$ and the $k$-th feature dimension is calculated as:
\begin{equation} 
s_{ij,k}=1-r_{ij,k}.
\end{equation} 
Propagation relationships with similar values of $s_{ij,k}$ tend to correspond to similar propagation patterns.

Importantly, $s_{ij,k}$ depends only on the relative feature difference between the connected nodes, rather than on the specific semantics of the corresponding feature dimension. 
Even when feature dimensions carry completely different meanings across graph domains, different graphs can still exhibit similar relative feature differences, thereby resulting in similar propagation relationships.
In this way, propagation relationships can be modeled across graph domains in a semantics-independent manner.

After obtaining the propagation relationships for all edges and feature dimensions, we further extract representative propagation relationships from multiple pre-training graph domains. Specifically, for the $i$-th pre-training graph $G^{(i)}$, we denote all propagation relationships as:
\begin{equation} 
\mathcal{S}^{(i)} = \{s_{uv,k}^{(i)} | e_{uv}^{(i)} \in \mathcal{E}^{(i)} , k = 1,\dots,d        \}.
\end{equation}
We then aggregate the propagation relationships from all graphs in the pre-training graph collection $\mathcal{G}_{pre} = \{G^{(1)},G^{(2)},\dots,G^{(M)}\}$:
\begin{equation} 
\mathcal{S}_{\text{pre}} = \bigcup_{i=1}^M \mathcal{S}^{(i)}.
\end{equation}
Based on $\mathcal{S}_{\text{pre}}$, we apply clustering to obtain a set of representative propagation relationship prototypes:
\begin{equation} 
\mathcal{P} = \text{Cluster} \left(\mathcal{S}_{\text{pre}}\right) = \{p_1, p_2, \dots, p_K\},
\end{equation}
where $K$ denotes the number of propagation relationship prototypes, $p_c$ represents the $c$-th prototype, and $\text{Cluster}\left(\cdot\right)$ denotes the clustering operation. In this work, we adopt K-means clustering \cite{kmeans} to obtain these propagation relationship prototypes. Propagation relationships with similar values are assigned to the same prototype, such that each prototype represents a representative propagation relationship shared across different graph domains.

Furthermore, each propagation relationship prototype $p_c$ is associated with a learnable scalar $\alpha_c$ to characterize the propagation strength corresponding to the propagation relationship represented by this prototype. The learnable propagation strengths associated with all propagation relationship prototypes are denoted as:
\begin{equation} 
\boldsymbol{\alpha} = [\alpha_1,\alpha_2,\dots,\alpha_K] \in \mathbb{R}^{K}.
\end{equation}
Accordingly, the propagation relationship prototype bank can be defined as:
\begin{equation} 
\mathcal{B} = \{(p_c, \alpha_c) \mid c = 1, \dots, K\}
\end{equation}
where each pair $(p_c, \alpha_c)$ consists of a propagation relationship prototype and its associated learnable propagation strength.

By organizing representative propagation relationships together with their corresponding propagation strengths, the prototype bank provides a unified and transferable representation of cross-domain propagation knowledge. It enables subsequent message passing to assign appropriate propagation strengths to different edges and feature dimensions according to their propagation relationships, thereby guiding feature propagation during message passing.

\subsection{Propagation-aware Message Passing}
Based on the propagation relationship prototype bank, we further introduce propagation-aware message passing to modulate information propagation across different edges and feature dimensions according to their propagation relationships.

ProGFM first transforms node feature representations into the message space of the current layer. 
Specifically, at the $l$-th layer, the feature representations $\mathbf{H}^{(l)}$ ($\mathbf{H}^{(0)}=\tilde{\mathbf{X}}$) are transformed as:
\begin{equation} 
\tilde{\mathbf{H}}^{(l)} = \mathbf{H}^{(l)} \mathbf{W}^{(l)},
\end{equation}
where $\mathbf{W}^{(l)}$ denotes the learnable transformation matrix, and $\tilde{\mathbf{H}}^{(l)}$ denotes the transformed feature representations at the $l$-th layer.

Based on the transformed feature representations, ProGFM performs propagation relationship matching for each edge and each feature dimension. Specifically, we frist calculate the relative feature difference for each edge and each feature dimension. For edge $e_{ij}$ and the $k$-th feature dimension, the relative feature difference is calculated as:
\begin{equation} 
r_{ij,k}^{(l)} = \frac{\left|\tilde{\mathbf{h}}_{i,k}^{(l)} - \tilde{\mathbf{h}}_{j,k}^{(l)} \right|}{\text{range}_k^{(l)}},
\end{equation}
\begin{equation} 
\text{range}_k^{(l)} = \text{max}(\tilde{\mathbf{H}}^{(l)}_{:.k}) - \text{min}(\tilde{\mathbf{H}}^{(l)}_{:.k}) + \epsilon.
\end{equation}
Accordingly, the propagation relationship between edge $e_{ij}$ and the $k$-th feature dimension can be expressed as:
\begin{equation} 
s_{ij,k}^{(l)} = 1-r_{ij,k}^{(l)}.
\end{equation}
For each propagation relationship $s_{ij,k}^{(l)}$, we match it with the nearest propagation relationship prototype in the prototype bank $\mathcal{B}$. The index of the matched prototype is calculated as:
\begin{equation} 
c_{ij,k}^{*(l)} = \arg \min_{c \in \{1, \dots, K\}} \left| s_{ij,k}^{(l)} - p_c \right|.
\end{equation}
Here, $c_{ij,k}^{*(l)}$ denotes the index of the propagation relationship prototype matched to edge $e_{ij}$ and the $k$-th feature dimension.
Since the $c$-th entry of the prototype bank is defined as $\mathcal{B}_c=\left( p_c,a_c\right)$, the matched propagation relationship prototype and its associated propagation strength are retrieved as:
\begin{equation} 
\left( p_{ij,k}^{*(l)}, z_{ij,k}^{(l)} \right) = \mathcal{B}_{c_{ij,k}^{*(l)}}.
\end{equation}
The propagation strengths corresponding to all feature dimensions are then combined to form an edge-level propagation strength vector:
\begin{equation} 
\mathbf{z}_{ij}^{(l)} = \left[ z_{ij,1}^{(l)}, z_{ij,2}^{(l)}, \dots, z_{ij,d}^{(l)} \right] \in \mathbb{R}^d.
\end{equation}

Based on the propagation strength vectors, ProGFM performs propagation-aware message passing by modulating the propagation of different feature dimensions. Specifically, the propagation-aware aggregation is formulated as:
\begin{equation} 
\mathbf{h}_i^{(l+1)} = \sigma \left( \sum_{j \in \mathcal{N}(i)} \hat{A}_{ij} \left[ \mathbf{z}_{ij}^{(l)} \odot \tilde{\mathbf{h}}_{j}^{(l)} \right] \right).
\end{equation}
where where $\mathbf{z}_{ij}^{(l)}$ denotes the propagation strength vector associated with edge $e_{ij}$, $\odot$ denotes element-wise multiplication, $\mathcal{N}(i)$ denotes the set of neighboring nodes of $v_i$, $\hat{A}_{ij}$ denotes the normalized adjacency coefficient, and $\sigma\left(\cdot\right)$ denotes the activation function.

Combining feature transformation, propagation relationship matching, and propagation-aware message passing, the overall process is formulated as:
\begin{equation} 
\mathbf{h}_i^{(l+1)} = \sigma \left( \sum_{j \in \mathcal{N}(i)} \hat{A}_{ij} \left[ \mathbf{z}_{ij}^{(l)} \odot \left( \mathbf{h}_j^{(l)} \mathbf{W}^{(l)} \right) \right] \right).
\end{equation}
In this way, ProGFM replaces the fixed message passing mechanism used by conventional GNNs with propagation-aware message passing, allowing each edge to modulate the propagation of different feature dimensions according to their propagation relationships.

\subsection{Self-supervised Pre-training and Zero-tuning Adaptation}
After $L$ layers of propagation-aware message passing, ProGFM obtains the final node representations, which are used for self-supervised pre-training and downstream tasks:
\begin{equation} 
\mathbf{H}^{(i,L)} = f_\theta \left( G^{(i)}, \mathcal{B} \right),
\end{equation}
where $\mathbf{H}^{(i,L)}$ denotes the final node representations of the $i$-th graph, $f_\theta$ denotes the ProGFM encoder parameterized by $\theta$, with $\theta$ including the learnable transformation matrices $\left\{ W^{(l)} \right\}_{l=0}^{L-1}$, and $\mathcal{B}$ denotes the propagation relationship prototype bank.

ProGFM is independent of a specific self-supervised pre-training objective and can be integrated with various graph self-supervised learning methods. During multi-domain pre-training, the ProGFM encoder and the learnable propagation strengths within $\mathcal{B}$ are jointly optimized across all graphs in $\mathcal{G}_\text{pre}$:
\begin{equation} 
\min_{\theta, \alpha} \sum_{G^{(i)} \in \mathcal{G}_{\text{pre}}} \mathcal{L}_{\text{ssl}} \left( \mathbf{H}^{(i,L)}, G^{(i)} \right).
\end{equation}
where $\mathcal{G}_\text{pre}$ denotes the set of pre-training graphs, $\boldsymbol{\alpha}$ denotes the learnable propagation strengths within $\mathcal{B}$, and $\mathcal{L}_{\text{ssl}}$ denotes the self-supervised pre-training objective. Following existing work \cite{TIG, TFSGFM}, we employ the structure-based self-supervised objective proposed in SGRL \cite{SGRL}, while other graph self-supervised objectives can also be readily incorporated.

\begin{table*}[htbp]
\caption{One-shot node classification performance. The best results are highlighted in bold, and the second-best results are underlined.}
\label{tb:node}
\centering
\resizebox{0.83\textwidth}{!}{
\begin{tabular}{lcccccc}
\toprule
\textbf{Method}   & \textbf{Cora} & \textbf{CiteSeer} & \textbf{PubMed} & \textbf{Photo} & \textbf{Computers} & \textbf{CS}\\ 
\midrule
GCN                         & 29.05 ± 7.84  & 28.57 ± 6.32 & 43.57 ± 8.03   & 46.43 ± 10.41  & 36.30 ± 12.84  & 60.00 ± 9.57\\ 
GAT                         & 27.39 ± 5.68  & 27.15 ± 5.00 & 44.15 ± 7.60   & 44.17 ± 11.48  & 34.99 ± 12.42  & 58.54 ± 8.78\\ 
\midrule
DGI                          & 30.56 ± 6.97  & 31.78 ± 6.17 & 44.70 ± 6.20   & 47.83 ± 8.00  & 37.37 ± 7.98  & 65.21 ± 6.93\\ 
BGRL                      & 36.28 ± 7.14  & 35.08 ± 6.80 & 43.98 ± 6.55   & 50.87 ± 8.78  & 40.82 ± 9.19 & 63.73 ± 6.16\\
GraphMAE                     & 40.59 ± 7.24  & 38.58 ± 7.25 & 49.09 ± 9.23   & 55.11 ± 9.59  & 46.73 ± 10.89  & 65.39 ± 6.68\\
\midrule
MDGFM                   & 43.42 ± 7.60  & 35.60 ± 7.01 & 46.17 ± 8.89   & 55.22 ± 9.01  & 46.08 ± 9.11  & 66.03 ± 7.40\\
MDGPT                    & 42.68 ± 7.30 & 43.80 ± 9.52 & 51.58 ± 9.52   & \underline{67.96 ± 9.58} & \underline{52.98 ± 10.74}  & 68.47 ± 8.01\\
SAMGPT                    & 43.49 ± 7.56 & 37.62 ± 7.31 & 44.00 ± 8.62   & 55.20 ± 8.35 & 47.38 ± 8.88 & 64.26 ± 7.56\\
\midrule
TIG                    & \underline{47.04 ± 9.52} & \textbf{45.50 ± 8.76} & \underline{51.79 ± 9.98}   & 63.22 ± 10.51 & 49.01 ± 11.48  & \underline{72.39 ± 7.12}\\
ProGFM                 & \textbf{49.25 ± 7.85}  & \underline{44.64 ± 8.56} & \textbf{52.45 ± 7.76}  & \textbf{70.47 ± 8.92} & \textbf{54.88 ± 9.92} & \textbf{75.28 ± 6.70}  \\
\bottomrule
\end{tabular}
}

\end{table*}

\begin{table*}[htbp]
\caption{One-shot subgraph classification performance. The best results are highlighted in bold, and the second-best results are underlined. Methods with “*” are reported from \cite{LEAD}.}
\label{tb:subgraph}
\centering
\resizebox{0.83\textwidth}{!}{
\begin{tabular}{lcccccc}
\toprule
\textbf{Method}   & \textbf{Cora} & \textbf{CiteSeer} & \textbf{PubMed} & \textbf{Photo} & \textbf{Computers} & \textbf{CS}\\ 
\midrule
DGI                          & 42.37 ± 09.32  & 34.29 ± 07.43 & 50.32 ± 09.44   & 58.20 ± 09.45  & 47.11 ± 10.09  & 67.67 ± 08.48\\ 
BGRL                      & 44.98 ± 07.25  & 37.87 ± 07.69 & 47.57 ± 08.49   & 59.71 ± 09.85  & 48.04 ± 09.42 & 65.49 ± 06.65\\
GraphMAE                     & 46.02 ± 08.45  & 42.21 ± 08.56 & 51.06 ± 09.97   & 61.96 ± 09.90  & 50.83 ± 10.60  & 68.14 ± 07.79\\
\midrule
MDGFM                   & 47.84 ± 8.15  & 38.04 ± 7.48 & 49.56 ± 9.87   & 60.74 ± 9.49  & 50.83 ± 10.25  & 69.06 ± 7.58\\
MDGPT                    & 45.88 ± 7.78 & 45.43 ± 9.93 & 52.13 ± 10.08   & \textbf{68.51 ± 9.59} & \underline{54.36 ± 11.16}  & \underline{69.29 ± 7.87}\\
SAMGPT                    & 49.18 ± 8.22 & 41.12 ± 8.40 & 48.19 ± 9.79   & 62.28 ± 9.13 & 54.06 ± 10.11 & 67.53 ± 7.75\\
\midrule
LEDA*                   & 50.70 ± 10.67 & 43.83 ± 10.04 & \underline{54.34 ± 11.08}   & 64.35 ± 9.46 & 51.00 ± 12.19  & 68.74 ± 8.00\\
TIG                    & \underline{52.26 ± 10.19} & \underline{46.16 ± 9.24} & 51.93 ± 11.21   & 63.06 ± 10.34 & 49.92 ± 11.85  & 68.08 ± 7.76\\
ProGFM                 & \textbf{56.36 ± 9.10}  & \textbf{46.54 ± 9.17} & \textbf{55.04 ± 9.67}  & \underline{68.20 ± 9.47} & \textbf{56.26 ± 11.81} & \textbf{73.82 ± 7.32}  \\
\bottomrule
\end{tabular}
}

\end{table*}

For downstream tasks, to evaluate the direct transferability of the learned cross-domain propagation knowledge, we adopt a zero-tuning adaptation strategy. Specifically, the pre-trained ProGFM encoder and propagation relationship prototype bank are directly transferred to an unseen target graph domain and remain frozen throughout downstream adaptation. Without applying any fine-tuning or prompt-based methods to the pre-trained model, ProGFM obtains the target-domain node representations as:
\begin{equation} 
\mathbf{H}^{(t,L)} = f_{\theta^*} \left( G^{(t)}, \mathcal{B}^* \right),
\end{equation}
where $G^{(t)}$ denotes the target graph, while $\theta^*$ and $\mathcal{B}^*$ denote the encoder parameters and propagation relationship prototype bank obtained from multi-domain pre-training, respectively.

Based on the node representations, ProGFM first constructs class prototypes using the labeled nodes in the target graph and then performs prototype-based classification. Specifically, for each class $q$, the corresponding class prototype is computed by averaging the node representations of labeled nodes belonging to this class:
\begin{equation} 
\mathbf{c}_q = \frac{1}{\left| V_q^{(t)} \right|} \sum_{v_i \in V_q^{(t)}} \mathbf{h}_i^{(t,L)},
\end{equation}
where $V_q^{(t)} \subseteq V^{(t)}$ denotes the set of labeled nodes belonging to class $q$ in the target graph, and $\mathbf{h}_i^{(t,L)}$ denotes the node representation of $v_i$. For each query node $v_i$, ProGFM computes the similarity between its node representation and all class prototypes, and assign the class with the highest similarity score as its predicted label:
\begin{equation} 
\hat{y}_i = \arg \max_{q \in \mathcal{C}^{(t)}} \text{sim} \left( \mathbf{h}_i^{(t,L)}, \mathbf{c}_q \right),
\end{equation}
where $\mathcal{C}^{(t)}$ denotes the class set of the target graph and $\text{sim}\left( \cdot, \cdot\right)$ denotes the similarity function.

In this way, ProGFM can directly transfer the propagation knowledge learned from multiple pre-training graph domains to unseen graph domains without modifying the pre-trained model.

\section{Experiments}
In this section, we conduct extensive experiments on node classification, subgraph classification, and graph classification tasks to evaluate the cross-domain generalization ability of ProGFM and verify whether the learned propagation knowledge can be directly transferred to unseen graph domains.

\subsection{Datasets}
For node classification and subgraph classification, we select six widely used node-level graph datasets, including the citation networks Cora, CiteSeer, and PubMed \cite{data_Cora1, data_Cora2}, the co-purchase networks Photo and Computers \cite{DataAmazon}, and the co-authorship network CS \cite{DataCS}. These datasets exhibit substantial differences in feature dimensionality, feature semantics, and graph structures. For graph classification, we adopt four widely used graph-level datasets from diverse domains, including social network classification datasets IMDB-BINARY and COLLAB \cite{IMDB}, as well as protein graph classification datasets PROTEINS and DD \cite{Proteins, DD, Proteins_DD}. The detailed statistics of these datasets are provided in Appendix \ref{ap:datasets}.

\subsection{Baselines}
We compare ProGFM with representative methods from three categories. First, we include GCN \cite{GCN} and GAT \cite{GAT} as representative conventional graph neural network baselines. Second, we select three widely used graph self-supervised learning methods, including DGI \cite{DGI}, BGRL \cite{BGRL}, and GraphMAE \cite{GraphMAE}. Third, we compare ProGFM with recent graph foundation models and cross-domain graph learning methods, including MDGFM \cite{MDGFM}, TIG \cite{TIG}, MDGPT \cite{MDGPT}, SAMGPT \cite{SAMGPT}, LEDA \cite{LEAD}, TFSGFM \cite{TFSGFM}, and SCR \cite{SCR}.

\subsection{Experimental Setup}
In all experiments, each downstream target graph remains unseen during multi-domain pre-training. For node classification and subgraph classification, we adopt a leave-one-graph-out setting. When one graph dataset is selected as the downstream target graph, the remaining datasets are used for multi-domain pre-training. For example, in node classification experiments, when Cora is selected as the downstream target graph, the remaining datasets, including CiteSeer, PubMed, Photo, Computers, and CS, are used for multi-domain pre-training. Following MDGPT \cite{MDGPT} and TIG \cite{TIG}, we construct 500 few-shot tasks for each target dataset and report the average classification accuracy over all tasks. For graph classification, ProGFM is pre-trained on the six node-level graph datasets, while IMDB-BINARY, COLLAB, PROTEINS, and DD are exclusively used as unseen downstream target datasets. Following the experimental protocols of SCR \cite{SCR} and ProG \cite{ProG}, we report classification accuracy and Macro-F1 score as the evaluation metrics. Further details of the experimental setup are provided in Appendix \ref{ap:Experimental}.

\subsection{One-Shot Node and Subgraph Classification}
We first evaluate ProGFM on one-shot node classification and subgraph classification tasks to assess its transferability under extremely limited target-domain supervision. The experimental results are summarized in Tables \ref{tb:node} and \ref{tb:subgraph}. For node classification tasks, ProGFM achieves strong performance across all evaluated datasets, obtaining the best results on five out of six datasets and competitive performance on the remaining dataset. Compared with existing graph foundation models, ProGFM achieves notable improvements, especially on Cora, Computers, Photo, and CS datasets. This can be attributed to the fact that existing graph foundation models mainly alleviate cross-domain discrepancies through feature alignment and/or structure alignment. However, feature semantics and structural patterns may vary substantially across different graph domains. Mapping different domains into a shared space may therefore fail to fully preserve all transferable information, leading to inconsistent transfer performance across different datasets. This suggests that adopting alignment-based strategies alone may not be sufficient to capture transferable knowledge across graph domains. In contrast, ProGFM regards the propagation relationships between edges and feature dimensions as transferable knowledge units. Despite the variations in feature semantics and graph structures across different domains, similar propagation relationships can still exist across graphs. By explicitly modeling these transferable propagation relationships, ProGFM can learn shared propagation knowledge and enable effective knowledge transfer across different graph domains. Similar improvements are also observed in subgraph classification tasks, where ProGFM achieves superior performance on most datasets. The consistent performance across node and subgraph classification tasks further validates the effectiveness of ProGFM in capturing transferable graph knowledge across different graph domains.

\begin{table*}[htbp]
\caption{Few-shot node classification performance. The best results are highlighted in bold, and the second-best results are underlined.}
\label{tb:few-shot}
\centering
\resizebox{0.96\textwidth}{!}{
\begin{tabular}{ll|ccc|ccc}
\toprule
\multirow{2}{*}{\textbf{Setting}} & \multirow{2}{*}{\textbf{Method}}   &      & \textbf{3-shot}   &         &      & \textbf{5-shot}   &       \\ \cmidrule{3-8}
 &  & \textbf{Cora} & \textbf{CiteSeer} & \textbf{PubMed}  & \textbf{Cora} & \textbf{CiteSeer} & \textbf{PubMed}\\ 

\midrule
\multirow{2}{*}{Prompt-tuning}      &MDGPT   & 53.65 ± 6.71 & 54.76 ± 6.04 & \underline{60.31 ± 6.50}  & 58.90 ± 5.11 & 59.78 ± 4.12 & \underline{63.44 ± 5.55} \\
                                    &SAMGPT               & 58.36 ± 5.66 & 55.42 ± 5.90 & 57.86 ± 7.77  & 63.99 ± 3.94 & 61.25 ± 3.84 & 61.53 ± 6.15\\
\midrule
\multirow{2}{*}{Zero-tuning}      &TIG   & 57.45 ± 7.31 & 56.82 ± 5.52 & 56.98 ± 6.78  & 62.50 ± 4.42 & 60.53 ± 3.93 & 58.68 ± 5.56 \\
                                    &ProGFM               & \underline{61.75 ± 5.80} & \underline{57.84 ± 5.14} & 56.31 ± 5.13  & \underline{66.21 ± 3.75} & \underline{62.00 ± 3.02} & 59.07 ± 4.63\\
\midrule
\multirow{1}{*}{Classifier-tuning}       &ProGFM-tuning     & \textbf{63.10 ± 5.72} & \textbf{58.73 ± 5.24} & \textbf{60.39 ± 5.33} & \textbf{68.15 ± 3.24} & \textbf{63.07 ± 3.18} & \textbf{65.81 ± 4.77}\\
\bottomrule
\end{tabular}
}
\end{table*}

\begin{table*}[htbp]
\caption{Graph classification performance. The best results are highlighted in bold. Methods with “*” are reported from \cite{SCR}.}
\label{tb:Graph classification}
\centering
\resizebox{0.98\textwidth}{!}{
\begin{tabular}{l c c c c c c c c}
\toprule
\multirow{2}{*}{\textbf{Method}} & \multicolumn{2}{c}{\textbf{IMDB-BINARY}} & \multicolumn{2}{c}{\textbf{COLLAB}} & \multicolumn{2}{c}{\textbf{PROTEINS}}  & \multicolumn{2}{c}{\textbf{DD}}\\
\cmidrule(lr){2-3} \cmidrule(lr){4-5} \cmidrule(lr){6-7} \cmidrule(lr){8-9}
        & \textbf{Acc} & \textbf{F1} & \textbf{Acc} & \textbf{F1} & \textbf{Acc} & \textbf{F1} & \textbf{Acc} & \textbf{F1} \\
\midrule
GCN*     & 57.30 ± 0.98   & 54.62 ± 1.12  & 47.23 ± 0.61   & 41.10 ± 0.39  & 56.36 ± 7.97   & 46.69 ± 10.82  & 55.33 ± 6.22   & 44.74 ± 4.23\\
\midrule

SCR*        & 61.83 ± 1.60   & 60.91 ± 2.18  & 65.45 ± 1.05   & 57.71 ± 1.82  & 68.54 ± 1.47   & 65.23 ± 1.37  & 69.96 ± 0.74   & 69.85 ± 0.51\\
TFSGFM      & 63.61 ± 1.33   & 63.20 ± 1.38  & 42.18 ± 2.66   & 42.27 ± 2.52  & 66.55 ± 1.98   & 65.25 ± 1.66  & 73.54 ± 0.74   & 72.88 ± 0.67\\
TIG         & 67.58 ± 2.07   & 67.32 ± 2.18  & 67.53 ± 0.83   & 66.31 ± 0.78  & 58.76 ± 3.92   & 57.74 ± 3.97  & 63.26 ± 2.03   & 62.24 ± 1.95\\
FLT         & \textbf{69.14 ± 1.63}   & \textbf{68.87 ± 1.96}  & \textbf{68.04 ± 0.87}   & \textbf{66.79 ± 0.91}   & \textbf{72.31 ± 0.67}    & \textbf{70.33 ± 1.02}   & \textbf{75.18 ± 0.74}   & \textbf{74.01 ± 0.85}\\
\bottomrule

\end{tabular}
}

\end{table*}

\subsection{Few-Shot Node Classification}

We further evaluate ProGFM under 3-shot and 5-shot node classification settings, where more target-domain labels are available for adaptation. The experimental results are summarized in Table \ref{tb:few-shot}. As more target-domain labels are provided, ProGFM consistently maintains strong performance under both 3-shot and 5-shot settings. Notably, even with the zero-tuning strategy, ProGFM still achieves competitive performance compared with existing prompt-based graph foundation models, including MDGPT and SAMGPT. Furthermore, we introduce a ProGFM-tuning variant, where all parameters of the pre-trained ProGFM encoder and propagation relationship prototype bank remain frozen, and only a lightweight classifier is optimized using the few labeled target-domain samples. Under this setting, ProGFM achieves further improvements and obtains the best performance across all evaluated datasets. These results further validate that propagation relationships between edges and feature dimensions can serve as transferable knowledge units across graph domains, as the propagation relationship prototype bank obtained from multi-domain pre-training can be directly applied to unseen graph domains without further adaptation.

\subsection{Graph Classification}
We further evaluate ProGFM on graph classification tasks to investigate whether the learned propagation knowledge can generalize from node-level graph domains to graph-level prediction scenarios. The experimental results are summarized in Table \ref{tb:Graph classification}. This setting represents a more challenging cross-domain scenario, where the downstream graph classification datasets exhibit larger domain discrepancies from the node-level graphs used during pre-training. In particular, PROTEINS and DD are protein-related graph datasets, which exhibit substantial domain differences from the pre-training datasets. Alignment-based graph foundation models such as TIG achieve competitive performance on IMDB-BINARY and COLLAB, where the domain discrepancy is relatively smaller. However, their performance decreases on PROTEINS and DD, indicating that alignment-based strategies may become less effective when the target domains exhibit larger distribution shifts. This suggests that relying solely on alignment-based strategies may not fully capture transferable knowledge across diverse graph domains. In contrast, ProGFM achieves the best performance on all evaluated graph classification datasets in terms of both accuracy and Macro-F1 score, outperforming existing graph foundation models. These results demonstrate that the propagation knowledge learned from node-level graph domains can be effectively transferred to graph-level tasks. By modeling propagation relationships between edges and feature dimensions, ProGFM captures transferable graph knowledge that is independent of specific feature semantics and graph structures, further validating that propagation relationships can serve as transferable knowledge units across different graph tasks and domains.

\subsection{Ablation Study}
\begin{figure}[htbp]
    \centering
    \includegraphics[width=0.45\textwidth]{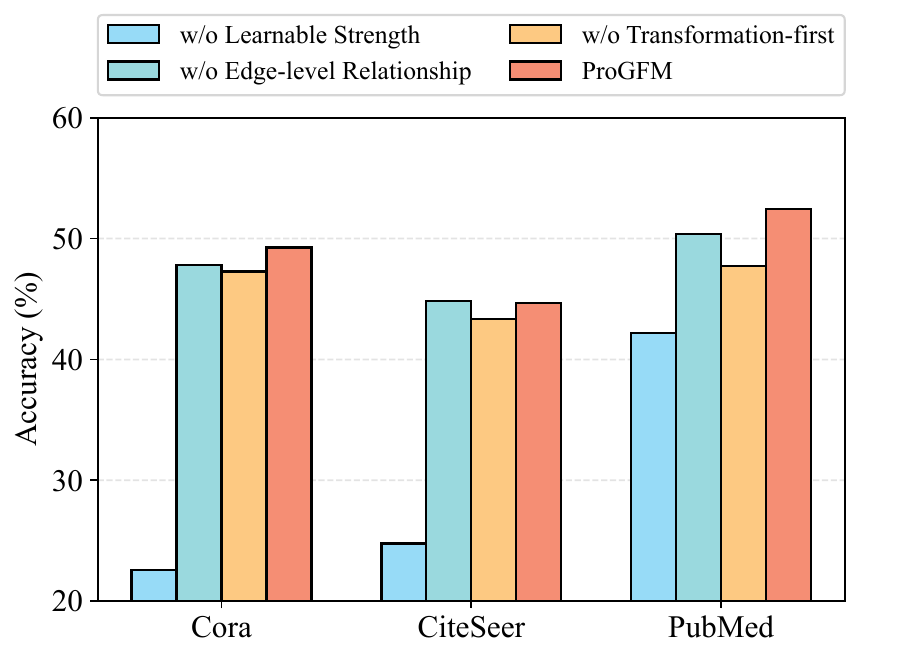}
    
    \caption{Ablation study results of ProGFM. The effectiveness of each key component is evaluated by comparing ProGFM with three variants.}
    \Description{Ablation study results of ProGFM. The effectiveness of each key component is evaluated by comparing ProGFM with three variants.}
    \label{fig:ablation}
\end{figure}

To investigate the effectiveness of the key designs in ProGFM, we conduct ablation studies with three variants, and the experimental results are summarized in Figure \ref{fig:ablation}. Specifically, we examine the effects of learnable propagation strengths, edge-level propagation relationships, and the propagation-aware message passing strategy. The three variants are denoted as w/o Learnable Strength, w/o Edge-level Relationship, and w/o Transformation-first, respectively. The w/o Learnable Strength variant replaces the learnable propagation strengths $\boldsymbol{\alpha}$ in the propagation relationship prototype bank $\mathcal{B}$ with fixed values initialized from a normal distribution $\mathcal{N}\left(1,1\right)$ and keeps them unchanged during pre-training. The w/o Edge-level Relationship variant replaces the edge-level propagation relationship modeling with a domain-level propagation relationship modeling strategy, where all edges within the same graph domain share identical propagation relationships for each feature dimension. The w/o Transformation-first variant changes the propagation-aware message passing strategy by performing propagation relationship matching and message passing before feature transformation.

As shown in Figure \ref{fig:ablation}, the variant without the learnable propagation strengths exhibits degraded performance compared with ProGFM. The performance drop indicates that different propagation relationship prototypes correspond to different propagation patterns, and their associated propagation strengths need to be adaptively learned. By optimizing the learnable propagation strengths $\boldsymbol{\alpha}$, ProGFM can automatically adjust the influence of different propagation relationship prototypes during message passing, leading to more effective cross-domain knowledge transfer.

As shown in Figure \ref{fig:ablation}, the w/o Edge-level Relationship variant suffers from performance degradation compared with ProGFM. This demonstrates that sharing the same propagation relationship among all edges within a graph domain is insufficient to capture their diverse propagation patterns. Even for the same feature dimension, different edges may correspond to different propagation relationships. By removing edge-level propagation relationship modeling, this variant fails to capture differences in propagation patterns among different edges within the same graph domain. These results demonstrate that modeling fine-grained propagation relationships between edges and feature dimensions is crucial for learning transferable graph knowledge.

As shown in Figure \ref{fig:ablation}, the w/o Transformation-first variant also suffers from performance degradation compared with ProGFM. In ProGFM, feature transformation is performed before propagation-aware message passing, ensuring that prototype matching and propagation strength modulation are applied in the message space of the current layer. When propagation-aware message passing is performed before feature transformation, the subsequent feature transformation mixes information across feature dimensions, thereby weakening the effect of propagation strength modulation.

\subsection{Parameter Sensitivity Analysis}
\begin{figure}[htbp]
    \centering
    \includegraphics[width=0.477\textwidth]{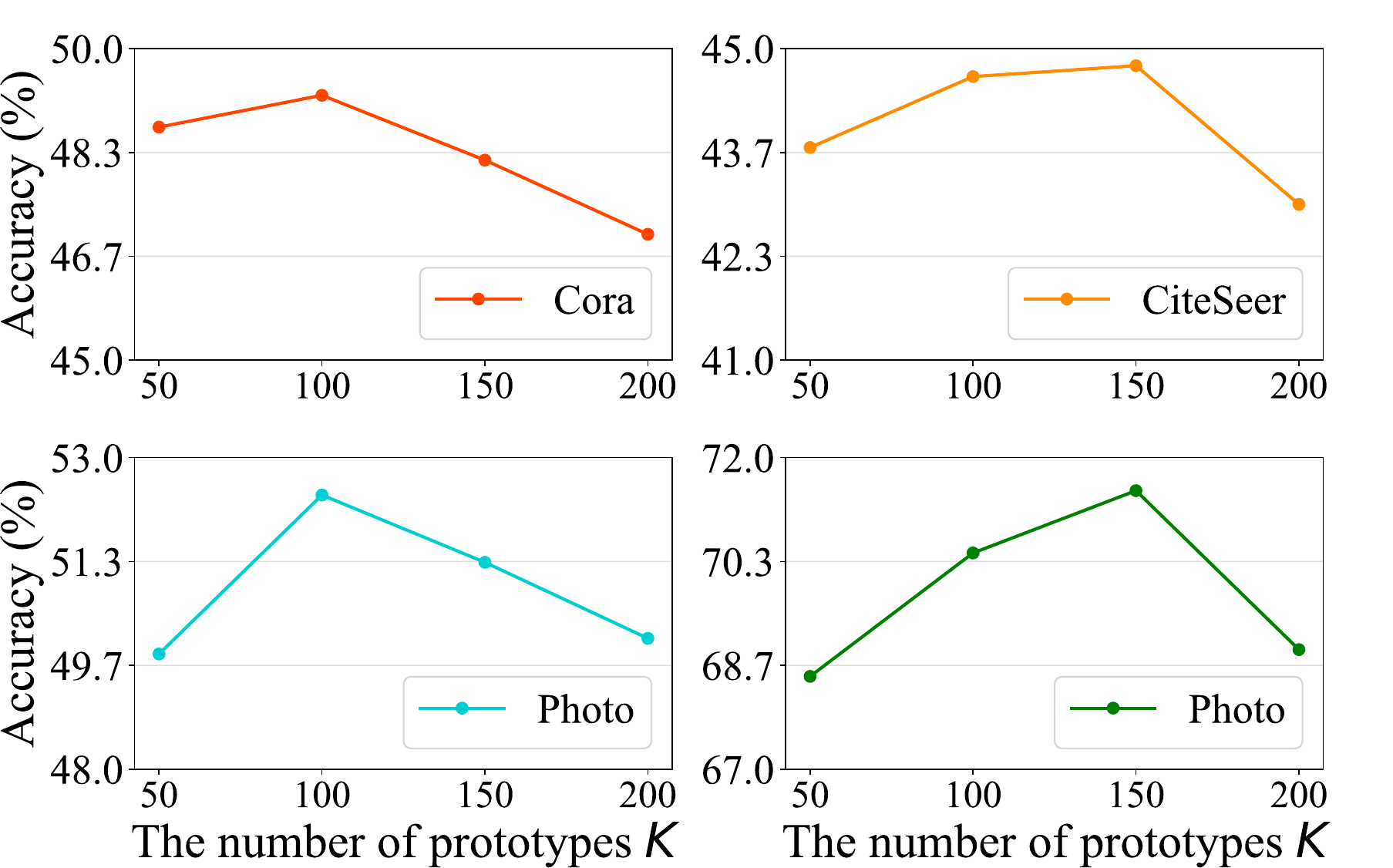}
    
    \caption{Sensitivity analysis of the number of propagation relationship prototypes $K$.}
    \Description{Sensitivity analysis of the number of propagation relationship prototypes $K$.}
    \label{fig:sensitivity}
\end{figure}

We further conduct a parameter sensitivity analysis on the size of the propagation relationship prototype bank, i.e., the number of prototypes $K$. The results are illustrated in Figure \ref{fig:sensitivity}. As observed, the performance of ProGFM exhibits a trend of first increasing and then decreasing as $K$ increases. When $K$ is relatively small, the prototype bank contains only a limited number of propagation relationship prototypes, which restricts its ability to capture the diverse propagation relationships across graph domains. As $K$ increases, the prototype bank gains stronger representation capacity and can model more fine-grained propagation relationships, leading to improved performance. However, with an excessively large $K$, a large number of redundant propagation relationship prototypes are introduced into the prototype bank, making it more difficult to accurately optimize their corresponding propagation strengths and consequently leading to performance degradation. Based on the overall performance across different datasets, we set $K=100$ as the default configuration for ProGFM.

\section{Conclusion}
In this paper, we identify propagation relationships between edges and feature dimensions as transferable knowledge units for graph foundation models and propose a Propagation-aware Graph Foundation Model (ProGFM) to learn cross-domain transferable propagation knowledge. By constructing a propagation relationship prototype bank and introducing propagation-aware message passing, ProGFM can learn transferable propagation knowledge from multiple graph domains and enable adaptive message propagation in unseen graph domains. Extensive experiments on node classification, subgraph classification, and graph classification tasks demonstrate that ProGFM achieves superior cross-domain generalization performance compared with existing graph foundation models. These results validate that propagation relationships can serve as transferable knowledge units across diverse graph tasks and domains. We hope that this work can inspire future research on discovering more fundamental and transferable knowledge underlying diverse graph domains.

\bibliographystyle{ACM-Reference-Format}
\bibliography{Mybib}

\appendix

\begin{table*}[htbp]
\caption{Dataset statistics.}
\label{aptb:data}
\centering
\resizebox{0.65\textwidth}{!}{
\begin{tabular}{lccccc} 
\toprule
\textbf{Dataset}  & \textbf{\#Graphs}   & \textbf{\#Avg.nodes}     & \textbf{\#Avg.edges}      & \textbf{\#Features}   & \textbf{\#Classes}   \\ 
\midrule
Cora    &1 & 2,708     & 10,556      & 1,433      & 7         \\
CiteSeer &1 & 3,327     & 9,104      & 3,703      & 6         \\
PubMed   &1 & 19,717    & 88,648     & 500        & 3         \\
Photo    &1 & 7,650     & 238,162    & 745        & 8         \\
Computers&1 & 13,752    & 491,722    & 767        & 10        \\ 
CS&1 & 18,333    & 163,788    & 6805        & 15        \\ 
\midrule
IMDB-BINARY     & 1,000     & 19.8          & 193.1         & -              & 2         \\
COLLAB     & 5,000     & 74.5          & 4,914.4         & -              & 3         \\
PROTEINS         & 1113     & 39.1         & 145.6         & 3              & 2         \\
DD              & 1,178    & 284.3        & 1431.3           & 89             & 2       \\

\bottomrule
\end{tabular}
}
\end{table*}

\begin{table*}[htbp]
\caption{Hyper-parameters settings of ProGFM.}
\label{tb:parameters}
\centering
\resizebox{0.85\textwidth}{!}{
\begin{tabular}{l c c c c  c c c c}
\toprule
\textbf{Scenario}  & \textbf{Lr} & \textbf{Weight\_decay} & \textbf{\#Epochs}  & \bm{$d$} & \bm{$K$}  & \bm{$L$} &\textbf{Dim \#1} & \textbf{Dim \#2}  \\

\midrule
\multirow{1}{*}{One-Shot Node Classification} 
                          &0.0001    &1e-4   &400     &16    &100  &2  &256    & 128 \\
                    
\midrule
\multirow{1}{*}{One-Shot Subgraph Classification} 
                          &0.0001    &1e-4   &400     &16    &100  &2  &256    & 128\\

\midrule
\multirow{1}{*}{Few-Shot Node Classification} 
                             &0.0001    &1e-4   &200    &16    &100  &2   &512  &512\\
\midrule
\multirow{1}{*}{Graph Classification} 
                            &0.00001    &1e-4   &300    &128    &100  &2  &512    & 256\\

\bottomrule

\end{tabular}
}
\end{table*}

\section{Dataset Statistics}
\label{ap:datasets}
The detailed statistics of these datasets are summarized in Table \ref{aptb:data}. It is worth noting that the IMDB-BINARY and COLLAB datasets do not provide raw node features; therefore, we construct node features based on node degree information.

\section{Detailed Experimental Setup}
\label{ap:Experimental}

\subsection{Implementation Details}
For node classification, ProGFM is first pre-trained on the source graph datasets and then directly transferred to the target graph. During downstream evaluation, all parameters of the pre-trained model are frozen, and the final-layer node representations are directly used for prototype-based classification. For subgraph classification, following FUG \cite{FUG} and LEDA \cite{LEAD}, we first obtain node representations using the pre-trained ProGFM encoder. Then, for each node, we construct its corresponding four-hop subgraph and perform structure-based parameter-free feature propagation over the node representations within the subgraph. The propagated representation of the center node is used as the final subgraph representation for prototype-based classification. For graph classification, ProGFM first generates node representations for each graph, and mean pooling is applied over all node representations to obtain the graph-level representation. Following SCR \cite{SCR} and ProG \cite{ProG}, 80\% of the graph samples in each target dataset are used as the test set, while a few labeled graphs are used to construct class prototypes for prototype-based classification.

\subsection{Hyper-parameter Settings}
To ensure a fair comparison, we adopts the same hyper-parameter settings across different target datasets within the same experiment. The detailed hyper-parameter settings are summarized in Table \ref{tb:parameters}, where $d$ denotes the unified feature dimensionality, $K$ denotes the number of propagation relationship prototypes in the prototype bank $\mathcal{B}$, $L$ denotes the number of propagation-aware message-passing layers, and Dim \#$l$ denotes the output dimensionality of the $l$-th feature transformation layer. The random seed is fixed to 0 for all experiments.

\section{Related Work}

\subsection{Graph Neural Networks}
Graph Neural Networks (GNNs) have become one of the most powerful approaches for graph representation learning by exploiting graph structures and node features through the message passing mechanism \cite{GNNsurvey2, GNNsurvey1}. By iteratively aggregating information from neighboring nodes, GNNs can effectively capture local structural dependencies and learn expressive representations for various graph tasks. GCN \cite{GCN} is a classic GNN model that learns node representations by directly aggregating information from neighboring nodes. GAT \cite{GAT} further introduces attention mechanisms to adaptively learn the importance of neighboring nodes during message passing. However, these methods are primarily designed for individual graph domains, where the learned representations are closely associated with domain-specific graph structures and feature distributions. Consequently, these models often cannot be directly applied to unseen graph domains and typically require additional adaptation or retraining, which limits their generalization ability across diverse graph environments.

\subsection{Graph Foundation Models}
Graph Foundation Models (GFMs) have recently emerged as a promising paradigm for improving the generalization ability of graph neural network \cite{GFMPosision}. Different from traditional GNNs that are usually designed for specific graphs and tasks, GFMs aim to learn generalizable knowledge from diverse graph domains and transfer it to various downstream scenarios \cite{BRIDGE, TFSGFM}.

To achieve cross-domain graph learning, existing GFMs mainly focus on reducing domain discrepancies through feature and structure alignment. From the perspective of feature alignment, UniGraph \cite{UniGraph} and GOFA \cite{GOFA} leverage large language models to encode textual node attributes from different domains into a unified representation space, facilitating knowledge transfer across different domains. MDGPT \cite{MDGPT} introduces domain tokens to capture domain-specific variations and map node features from different domains into a unified semantic space, while TIG \cite{TIG} reconstructs feature representations based on relative relationships among features to enhance feature consistency across domains. From the perspective of structure alignment, MDGFM \cite{MDGFM} and TFSGFM \cite{TFSGFM} employ graph structure learning methods to reconstruct edges based on node similarities, thereby aligning connectivity patterns across different graph domains. SAMGPT \cite{SAMGPT} proposes structure tokens to align structural distributions across multiple domains. Alternatively, some studies address structural discrepancies through architecture design. AnyGraph \cite{AnyGraph}, KDEM \cite{KDEM} and OMOG \cite{OMOG} adopt Mixture-of-Experts (MoE) architectures, where different experts are encouraged to capture distinct graph patterns and are adaptively selected for different graph domains, thereby alleviating structural discrepancies among diverse graphs. However, these approaches mainly alleviate distribution shifts across graph domains through feature and structure alignment, while overlooking the exploration of transferable knowledge units underlying graph data.

To address this limitation, a line of research has explored transferable knowledge units for graph foundation models from a structural perspective. GFT \cite{GFT} considers computation trees derived from message passing as transferable graph vocabulary, while RiemannGFM \cite{RiemannGFM} explores shared structural knowledge by constructing a structural vocabulary from common graph substructures, such as trees and cycles, and modeling them in Riemannian geometric spaces. These studies demonstrate the importance of identifying domain-shared structural patterns for improving graph generalization. However, graphs inherently consist of both structural information and node features \cite{GNNsurvey1}. Existing structural vocabularies mainly characterize graph structures independently, while failing to capture the interaction between structures and features. Exploring transferable knowledge units that capture the interaction between graph structures and node features can provide a promising direction for enabling more effective cross-domain knowledge transfer.

\end{document}